\documentclass[runningheads]{llncs}
\usepackage{graphicx}
\usepackage{comment}
\usepackage{amsmath,amssymb} 
\usepackage{color}

\usepackage{booktabs, subcaption, caption} 
\usepackage{multirow}
\usepackage{wrapfig,lipsum}
\usepackage{floatrow}
\newfloatcommand{capbtabbox}{table}[][\FBwidth]
\usepackage{url}

\begin{document}
\pagestyle{headings}
\mainmatter
\def\ECCVSubNumber{2257}  

\title{The Devil is in Classification: A Simple Framework for Long-tail Instance Segmentation}

\titlerunning{The Devil is in Classification}
%
\author{Tao Wang\inst{1,4}\orcidID{0000-0002-2480-878X} \and
Yu Li\inst{2,4} \and
Bingyi Kang\inst{4} \and
Junnan Li\inst{3} \and
Junhao Liew\inst{4} \and
Sheng Tang\inst{2} \and
Steven Hoi\inst{3} \and
Jiashi Feng\inst{4}
}
\authorrunning{T. Wang et al.}
%
\institute{NGS, National University of Singapore, Singapore
\email{twangnh@gmail.com}\and
Institute of Computing Technology, Chinese Academy of Sciences, China
\email{\{liyu,ts\}@ict.ac.cn} \and
Salesforce Research Asia, Singapore
\email{\{junnan.li,shoi\}@salesforce.com} \and
ECE Department, National University of Singapore, Singapore
\email{\{kang,liewjunhao\}@u.nus.edu} \email{elefjia@nus.edu.sg}}

\maketitle


\begin{abstract}
Most existing object instance detection and segmentation models only work well on fairly balanced benchmarks where per-category training sample numbers are comparable, such as COCO. They tend to suffer performance drop on realistic datasets that are usually long-tailed.  
This work aims to study and address such open challenges. 
Specifically, we systematically investigate performance drop of the state-of-the-art two-stage instance segmentation model Mask R-CNN on the recent long-tail LVIS dataset, and unveil that a major cause is the inaccurate classification of object proposals. 
Based on such an observation, we first consider various techniques for improving long-tail classification performance which indeed enhance instance segmentation results.
We then propose a simple calibration framework to more effectively alleviate classification head bias with a bi-level class balanced sampling approach. 
Without bells and whistles, it significantly boosts the performance of instance segmentation for tail classes on the recent LVIS dataset and our sampled COCO-LT dataset.
Our analysis provides useful insights for solving long-tail instance detection and segmentation problems, and the straightforward \emph{SimCal} method can serve as a simple but strong baseline. With the method we have won the 2019 LVIS challenge \footnote{\textbf{Importantly}, after the challenge submission~\cite{wang2019classification}, we find significant improvement can be further achieved by modifying the head from 2fc\_rand to 3fc\_ft (refer to Sec.~\ref{design_choice_calibration_head} and Table~\ref{htc_results}  for details), which is expected to generates much higher test set result. We also encourage readers to read our following work~\cite{li2020overcoming} that more effectively calibrates the last classification layer with a re-designed softmax module.}. Codes and models are available at \url{https://github.com/twangnh/SimCal}. 

\end{abstract}

\section{Introduction}

Object detection and instance segmentation
aim to localize and segment individual object instances from an input image. 
The widely adopted solutions to such tasks are built on region-based two-stage frameworks, \textit{e.g.}, Faster R-CNN~\cite{ren2015faster} and Mask R-CNN~\cite{he2017mask}.
Though these models have demonstrated remarkable performance on several class-balanced benchmarks, such as Pascal VOC~\cite{everingham2010pascal}, COCO~\cite{lin2014microsoft}
and OpenImage~\cite{openimage}, 
they are seldom evaluated on datasets with long-tail distribution that is common in realistic scenarios~\cite{reed2001pareto} and dataset creation~\cite{everingham2010pascal},\cite{krishna2017visual},\cite{lin2014microsoft}. 
Recently, Gupta et al.~\cite{gupta2019} introduce the LVIS dataset for large vocabulary long-tail instance segmentation model development and evaluation. 
They observe the long-tail distribution can lead to severe performance drop of the state-of-the-art instance segmentation model~\cite{gupta2019}. 
However, the reason for such performance drop is not clear yet.

\begin{figure}[!t]
	\centering
	\includegraphics[width=\textwidth]{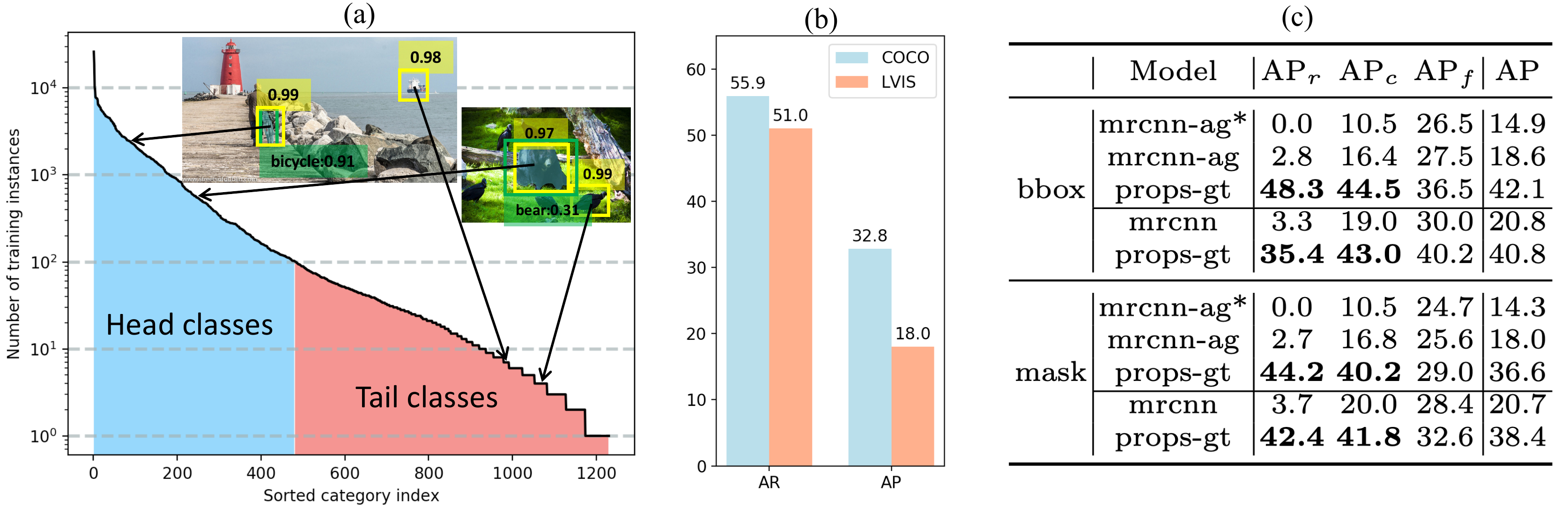}
	\caption{ (a) Examples of object proposal and instance segmentation results from ResNet50-FPN Mask R-CNN, trained on long-tail LVIS dataset. 
    The RPN can generate high-quality object proposals (yellow bounding boxes with high confidence scores) even on long-tail distribution, \textit{e.g.}, cargo ship (7 training instances) and vulture (4 training instances). 
    However, they are missed in final detection and segmentation outputs (green bounding boxes and masks) due to poor proposal classification performance. 
    Other proposal candidates and detection results are omitted from images for clarity. 
    (b) Comparison of proposal recall (COCO style Average Recall) and AP between COCO and LVIS dataset with Mask R-CNN model. 
    (c) Pilot experiment results on Mask R-CNN with class-agnostic and class-wise box and mask heads on ResNet50-FPN backbone, evaluated with LVIS v0.5 val set. 
    \emph{mrcnn-ag*} denotes standard inference with 0.05 confidence threshold as in optimal settings of COCO, while \emph{mrcnn-ag} means inference with threshold 0.0. Note for all later experiments we use 0.0 threshold. AP$^{bb}$ denotes box AP. \emph{props-gt} means testing with ground truth labels of the proposals}
	\label{fig_intro}
\end{figure}

In this work, we carefully study why existing models are challenged by long-tailed distribution and develop solutions accordingly. 
Through extensive analysis on Mask R-CNN in Sec.~\ref{pilot_analysis}, 
we show one major cause of performance drop is the inaccurate classification of object proposals,
which is referred to the bias of classification head.
Fig.~\ref{fig_intro} shows a qualitative example. 
Due to long-tail distribution, under standard training schemes, object instances from the tail classes are exposed much less frequently to the classifier than the ones from head classes\footnote{we use head classes and many-shot classes interchangeably.}, leading to poor classification performance on tail classes.  

To improve proposal classification, we first consider incorporating several common strategies developed for long-tail classification into current instance segmentation frameworks, including loss re-weighting~\cite{huang2016learning},\cite{tang2008svms}, adaptive loss adjustment (focal loss~\cite{lin2017focal}, class-aware margin loss~\cite{cao2019learning}), and data re-sampling~\cite{he2009learning,shen2016relay}. 
We find such strategies indeed improve long-tail instance segmentation performance, but their improvement on tail classes is limited and facing the trade-off problem of largely sacrificing performance on head classes.
We thus propose a simple and efficient framework after a thorough analysis of the above strategies.
Our method, termed \emph{SimCal}, aims to correct the bias in the classification head with a decoupled learning scheme. Specifically, after normal training of an instance segmentation model, it first collects class balanced proposal samples with a new bi-level sampling scheme that combines image-level and instance-level sampling, and then uses these collected proposals to calibrate the classification head. Thus performance on tail classes can be improved. \emph{SimCal} also incorporates a simple dual head inference component that effectively mitigates performance drop on head classes after calibration.

Based on our preliminary findings, extensive experiments are conducted on LVIS~\cite{gupta2019} dataset to verify the effectiveness of our methods.
We also validate the proposed method with SOTA multi-stage instance segmentation model  HTC~\cite{chen2019hybrid} and our sampled long-tail version of COCO dataset (COCO-LT). From our systematic study, we make the following intriguing observations:

\begin{itemize}
	\item Classification is the primary obstacle preventing state-of-the-art region-based object instance detection and segmentation models from working well on long-tail data distribution. There is still a large room for improvement along this direction.
	\item By simply calibrating the classification head of a trained model with a bi-level class balanced sampling in the decoupled learning scheme, the   performance for tail classes can be effectively improved.
	
\end{itemize}

\section{Related Works}

\subsubsection{Object Detection and Segmentation} 
Following the success of R-CNN~\cite{girshick2014rich}, Fast R-CNN~\cite{girshick2015fast} and Faster R-CNN~\cite{ren2015faster} architectures, the two-stage pipeline has become prevailing for object detection. 
Based on Faster R-CNN, He et al.~\cite{he2017mask} propose Mask R-CNN that extends the framework to instance segmentation with a mask prediction head to predict region based mask segments. 
Lots of later works try to improve the two-stage framework for object detection and instance segmentation. For example,~\cite{huang2019mask},\cite{jiang2018acquisition} add IOU prediction branch to improve confidence scoring for object detection and instance segmentation respectively. 
Feature augmentation and various training techniques are thoroughly examined by \cite{liu2018path}. 
Recently, \cite{cai2018cascade} and \cite{chen2019hybrid} further extend proposal based object detection and instance segmentation to multi-stage and achieve state-of-the-art performance. 
In this work, we study how to improve proposal-based instance segmentation models over long-tail distribution.

\subsubsection{Long-tailed Recognition}
Recognition on long-tail distribution is an important research topic as imbalanced data form a common obstacle in real-world applications. 
Two major approaches tackling long-tail problems are sampling~\cite{buda2018systematic},\cite{byrd2019effect}, \cite{he2009learning},\cite{shen2016relay} and loss re-weighting~\cite{cui2019class},\cite{huang2016learning},\cite{huang2019deep}. 
Sampling methods over-sample minority classes or under-sample majority classes to achieve data balance to some degree.  
Loss re-weighting assigns different weights to different classes or training instances adaptively, \textit{e.g.}, by inverse class frequency.
Recently, \cite{cui2019class} proposes to re-weight loss by the number of inversed effective samples. 
\cite{cao2019learning} explores class aware margin for classifier loss calculation. 
In addition,~\cite{kang2019decoupling} tries to examine the relation of feature and classifier learning in an imbalanced setting. \cite{wang2017learning} develops a meta learning framework that transfers knowledge from many-shot to few-shot classes. 
Existing works mainly focus on classification, while the crucial tasks of long-tail object detection and segmentation on remain largely unexplored. 

\section{Analysis: Performance Drop on Long-tail Distribution }
\label{pilot_analysis}

We investigate the performance decline phenomenon of  popular two-stage frameworks for long-tail instance detection and segmentation. 

Our analysis is based on experiments on LVIS v0.5 train and validation sets.
The LVIS dataset \cite{gupta2019} is divided into 3 sets: \emph{rare, common, and frequent}, among which \emph{rare} and \emph{common} contain tail classes and \emph{frequent} includes head classes. 
We report AP on each set, denoted as AP$_r$, AP$_c$, AP$_f$.
For simplicity, we train a baseline Mask R-CNN with ResNet50-FPN backbone and \emph{class agnostic box and mask prediction heads}.
As shown in Fig.~\ref{fig_intro} (c), our baseline model (denoted as mrcnn-ag*) performs poorly, especially on tail categories (\emph{rare} set, AP$_r$, AP$_r^{b}$), with 0 box and mask AP. 

Usually, the confidence threshold is set to a small positive value (\textit{e.g.} 0.05 for COCO) to filter out low-quality detections. 
Since LVIS contains 1,230 categories, the softmax activation gives much lower average confidence scores, thus we minish the threshold here.
However, even lowering the threshold to 0 (mrcnn-ag), the performance remains very low for tail classes, and improvement on \emph{rare} is much smaller than that of \emph{common} (6.1 vs 2.7 for segmentation AP, 5.9 vs 2.8 for bbox AP).
This reveals the Mask R-CNN model trained with the normal setting is heavily biased to the head classes. 

We then calculate proposal recall of \emph{mrcnn-ag} model and compare with the one trained on COCO dataset with the same setting.
As shown in Fig.~\ref{fig_intro} (b), the same baseline model trained on LVIS only has a drop of 8.8\% (55.9 to 51.0) in proposal recall compared with that on COCO, but notably, has a 45.1\% (32.8 to 18.0) drop in overall mask AP. 
Since the box and mask heads are class agnostic, we hypothesize that the performance drop is mainly caused by the degradation of proposal classification. 

To verify this, for the proposals generated by RPN~\cite{ren2015faster}, we assign their ground truth class labels to the second stage as its classification results. 
Then we evaluate the AP.
As shown in Fig.~\ref{fig_intro} (c), the mask AP for tail classes is increased by a large margin, especially on \emph{rare} and \emph{common} sets.
Such findings also hold for the box AP. 
Surprisingly, with normal class-wise box and mask heads (standard version of Mask R-CNN), performance on tail classes is also boosted significantly. This suggests the box and mask head learning are less sensitive to long-tail training data than classification. 

The above observations indicate that the low performance of the model over tail classes is mainly caused by poor proposal classification on them. We refer to this issue as \textit{classification head bias}. Addressing the bias is expected to effectively improve object detection and instance segmentation results.

\section{Solutions: Alleviating Classification Bias}
Based on the above analysis and findings, we first consider using several existing strategies of long-tail classification, and then present a new calibration framework to correct the classification bias for better detection and segmentation on long-tail distribution.

\subsection{Using Existing Long-tail Classification Approaches}
\label{other_approaches}
We adapt some popular approaches of image classification to solving our long-tail instance detection and segmentation problem, as introduced below. We conduct experiments to see how our adapted methods work in Sec. \ref{comp_other_methods}.
Given a sample $x_i$, the model outputs logits denoted as $y_i$, and $p_i$ is probability prediction on the true label $z$.

\subsubsection{Loss Re-weighting}~\cite{cui2019class}, \cite{huang2016learning}, \cite{khan2017cost}, \cite{tang2008svms}, \cite{ting2000comparative}
This line of works alleviate the bias by applying different weights to different samples or categories, such that tail classes or samples receive higher attention during training, thus improving the classification performance.
For LVIS, we consider a simple and effective inverse class frequency re-weighting strategy adopted in~\cite{huang2016learning,wang2017learning}. Concretely, the training samples of each class are weighted by $w=N/N_{j}$ where $N_{j}$ is the training instance number of class $j$. $N$ is a hyperparameter. To handle noise, the weights are clamped to $[0.1, 10.0]$. The weight for the background is also a hyperparameter. During training, the second stage classification loss is weighted as $L = -w_i\log(p_i)$.

\subsubsection{Focal Loss}~\cite{lin2017focal} Focal loss can be regarded as loss re-weighting that adaptively assigns a weight to each sample by the prediction. It was originally developed for foreground-background class imbalance for one-stage detectors, and also applicable to alleviating the bias in long-tail problems since head-class samples tend to get smaller losses due to sufficient training, and the influence of tail-class samples would be again enlarged.
Here we use the multi-class extension of Focal loss $L = -(1-p_i)^\gamma \log(p_i)$.

\subsubsection{Class-aware Margin Loss}~\cite{cao2019learning} 
This method assigns a class dependent margin to loss calculation. Specifically, a larger margin will be assigned for the tail classes, so they are expected to generalize better with limited training samples.
We adopt the margin formulation $\varDelta_j=C/N_{j}^{1/4}$~\cite{cao2019learning} where $N_{j}$ is the training instance number $N_{j}$ for class $j$ as above and plug the margin into cross entropy loss $L=-\log {e^{y_{iz}-\varDelta_z}}/ ({e^{y_{iz}-\varDelta_z}+\sum_{c\neq z}e^{y_{ic}}})$.

\subsubsection{Repeat Sampling}~\cite{he2009learning}, \cite{shen2016relay} Repeat sampling directly over-samples data (images) with a class-dependent repeating factor, so that the tail classes can be more frequently involved in optimization. 
Consequently, the training steps for each epoch will be increased due to the over-sampled instances.
However, this type of methods are not trivially applicable to detection frameworks since multiple instances from different classes frequently exist in one image. 
\cite{gupta2019} developed a specific sampling strategy for LVIS dataset, calculating a per-image repeat factor based on a per-category repeat threshold and over-sampling each training image according to the repeat factor in each epoch. 
Note that box and mask learning will also be affected by this method.

We implement the adapted version of ~\cite{cao2019learning}, \cite{cui2019class}, \cite{gupta2019}, \cite{lin2018focal} for experiments. See  Sec. \ref{comp_other_methods} for details. From the results, we find the above approaches indeed bring some performance improvements over the baselines, which however are very limited. 
\textbf{Re-weighting methods} tend to complicate the optimization of deep models with extreme data imbalance~\cite{cui2019class}, which is the case for object detection with long-tail distribution, leading to poor performance on head classes. 
\textbf{Focal loss} well addresses the imbalance between foreground and easy background samples, but  has difficulty in tackling the imbalance between foreground object classes with more similarity and correlation.
For \textbf{class-aware margin loss}, the prior margin enforced in loss calculation also complicates the optimization of a deep model, leading to larger drop of performance on head classes.
The \textbf{repeat sampling} strategy suffers from overfitting since it repeatedly samples from tail classes. Also, it additionally samples more data during training, leading to increased computation cost.
In general, the diverse object scale and surrounding context in object instance detection further complicate above-discussed limitations, making these methods hardly suitable for our detection tasks.

\subsection{Proposed \emph{SimCal}: Calibrating the Classifier}

\begin{figure}[!t]
	\centering
	\includegraphics[width=\linewidth]{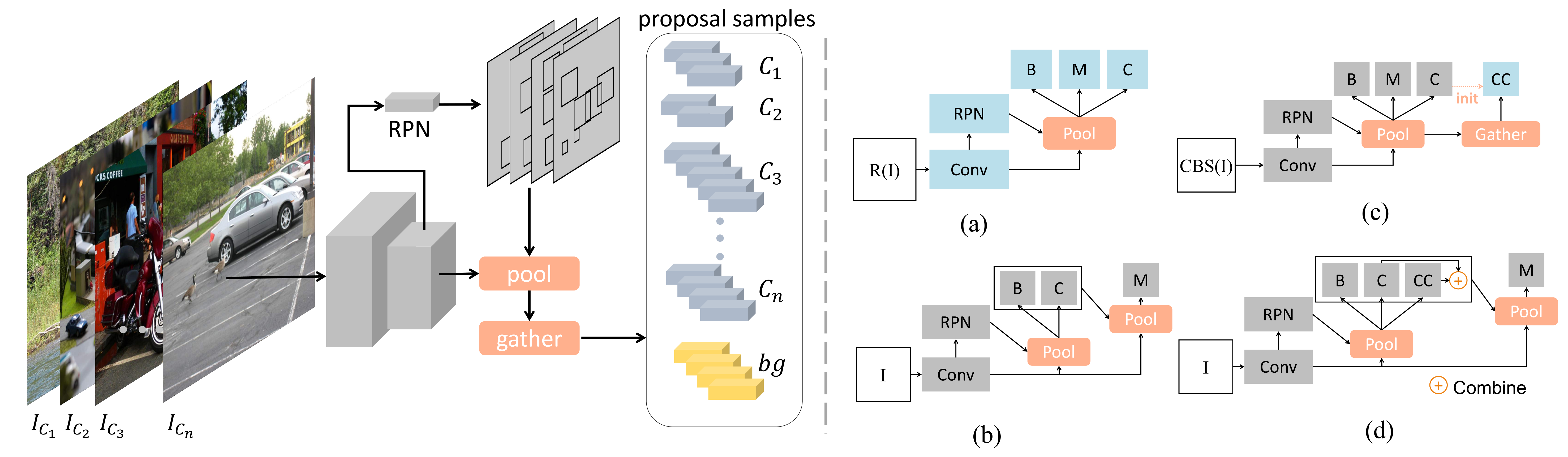}
	\caption{ Left: Illustration of proposed bi-level sampling scheme. Refer to Sec.~\ref{bi_level_sampling_calibration} for more details. Right: Architecture of proposed method. I: training or test image sets; R: random sampling; CBS: class-balanced sampling; C: classification head; B: box regression head; M: mask prediction head; CC: calibrated classification head. \emph{Blue} modules are in training mode while \emph{grey} modules indicate frozen. (a) (b) show standard Mask R-CNN training and inference respectively. (c) (d) show proposed calibration and dual head inference respectively
	}
	\label{bi_level_sampling_and_arch_diagram}
\end{figure}

We find in Sec.~\ref{pilot_analysis} that significant performance gain on tail classes can be achieved with merely ground truth proposal labels, and as discussed in Sec. \ref{other_approaches}, exiting classification approaches are not very suitable for tackling our long-tail instance segmentation task. 
Here, we propose a new \emph{SimCal} framework to calibrate the classification head by retraining it with a new bi-level sampling scheme while keeping the other parts frozen after standard training. 
This approach is very simple and incurs negligible additional computation cost since only the classification head requires gradient back-propagation. The details are given as follows.

\subsubsection{Calibration Training with Bi-level Sampling}
\label{bi_level_sampling_calibration}

As shown in Fig.~\ref{bi_level_sampling_and_arch_diagram}, we propose a bi-level sampling scheme to collect training instances for calibrating the classification head through retraining. 
To create a batch of training data, first, $n$ object classes (\textit{i.e.}, $c_1$ to $c_n$) are sampled uniformly form all the classes (which share the same probability).
Then, we randomly sample images that contain the categories respectively (\textit{i.e.}, I$_{c_1}$ to I$_{c_n}$), and feed them to the model. At the object level, we only collect proposals that belong to the sampled classes and background for training. 
Above, we only sample 1 image for each sampled class for simplicity, but note that the number of sampled images can also be larger.
As shown in Fig.~\ref{bi_level_sampling_and_arch_diagram} right (a), after standard training, we freeze all the model parts (including backbone, RPN, box and mask heads) except for the classification head, and employ the bi-level sampling to retrain the classification head, which is initialized with the original head. 
Then, the classification head is fed with fairly balanced proposal instances, thus enabling the model to alleviate the bias. 
Different from conventional fine-tuning conducted on a small scale dataset after pretraining on a large one, our method only changes the data sample distribution. 
Refer to supplementary material for more implementation details, including foreground and background ratio and matching IOU threshold for proposals.
Formally, the classification head is trained with loss:
\begin{equation}
\begin{aligned}
L = \frac{1}{\sum_{i=0}^{N}n_{i}}\sum_{i=0}^{N} \sum_{j=1}^{n_{i}}L_{cls}(p_{ij},p^{*}_{ij})
\end{aligned}
\label{cal_loss}
\end{equation}
where $N$ is the number of sampled classes per batch, $n_{i}$ is the number of proposal samples for class $i$, $i=0$ is for background, $L_{cls}$ is cross entropy loss, and $p_{ij}$ and $p^{*}_{ij}$ denotes model prediction and ground truth label.

\subsubsection{Dual Head Inference}
\label{dual_head}
After the above calibration, the classification head is now balanced over classes and can perform better on tail classes.
However, the performance on head classes drops. 
To achieve optimal overall performance, 
here we consider combining the new balanced head and the original one that have higher performance respectively on tail classes and on head classes. We thus propose a dual head inference architecture.

An effective combining scheme is to simply average the models' classification predictions~\cite{alpaydin1993multiple}, \cite{breiman1996bagging}, \cite{krogh1995neural},
but we find this is not optimal as the original head is heavily biased to many-shot classes. 
Since the detection models adopt class-wise post-processing (\textit{i.e.}, NMS) and the prediction does not need to be normalized, we propose a new combining scheme that directly selects prediction from the two classifiers for the head and tail classes:
\begin{equation}
p[z]=
\begin{cases}
p^{cal}[z] &N_{z}\leq T\\
p^{orig}[z] &\text{otherwise},
\end{cases}
\label{concat}
\end{equation}
where $z \in [0,C]$ indexes the classes, $C$ is the number of classes, $z=0$ stands for background, $p^{cal}$ and $p^{orig}$ denote the ($C$+$1$)-dimensional predictions of calibrated and original heads respectively, $p$ is the combined prediction, $N_{z}$ is the training instance number of class $z$, and $T$ is the threshold number controlling the boundary of head and tail classes.
Other parts of inference remain the same  (Fig.~\ref{bi_level_sampling_and_arch_diagram} (d)).
Our dual head inference is with small overhead compared to the original model.

\subsubsection{Bi-level Sampling vs. Image Level Repeat Sampling}
Image level repeat sampling (\textit{e.g.}, ~\cite{gupta2019}), which is traditionally adopted, balances the long-tail distribution at the image level, while our bi-level sampling alleviates the imbalance at the proposal level.  
Image level sampling approaches train the whole model directly, while we decouple feature and classification head learning, and  adjust the classification head only with bi-level class-centric sampling and keep other parts the freezed after training under normal image-centric sampling.
We also empirically find the best setting (t=0.001) of IS~\cite{gupta2019} additionally samples about 23k training images (56k in total) per epoch, leading to more than 40\% increase of training time. Comparatively, our method incurs less than 5\% additional time and costs much less GPU memory since only a small part of the model needs backpropagation.

\section{Experiments}
In this section, we first report experiments of using exiting classification approaches to solve our long-tail instance segmentation problem. Then we evaluate our proposed solution, \textit{i.e.} the \emph{SimCal} framework, analyze its model designs and test its generalizality.  

Our experiments are mainly conducted on  LVIS dataset~\cite{gupta2019}. Besides, to check the generalizability of our method, we sample a new COCO-LT dataset from COCO~\cite{lin2014coco}. We devise a complimentary instance-centric category division scheme that helps to more comprehensively analyze model performance. For each experiment, we report result with median overall AP over 3 runs.

\subsection{Datasets and Metrics}

\begin{wrapfigure}[14]{r}{0.5\textwidth}
\centering
\includegraphics[width=0.8\textwidth]{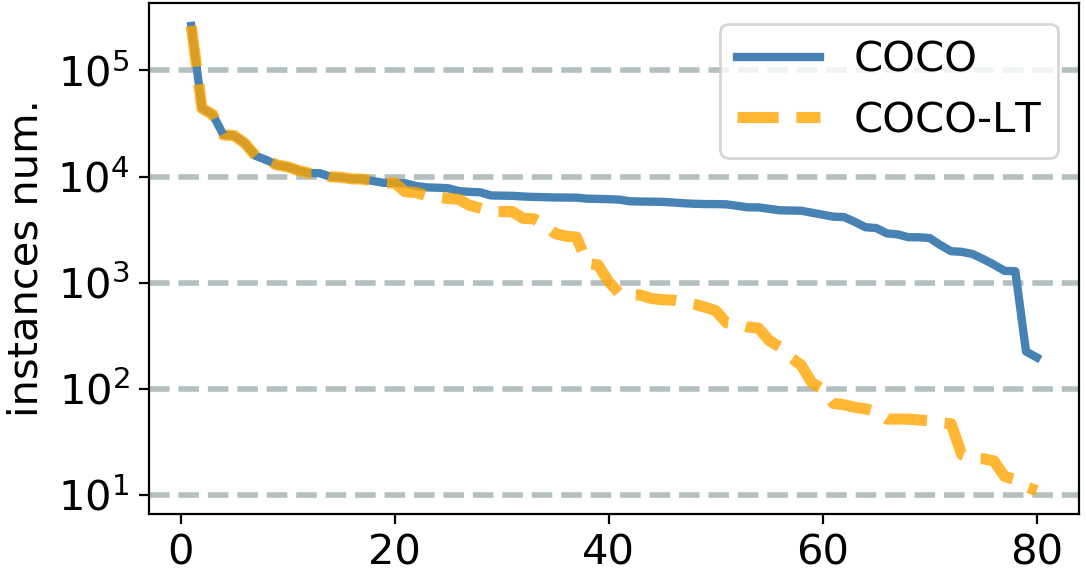}
\caption{ Category distribution of COCO (2017) and sampled COCO-LT datasets. The categories are sorted in descending numbers of training instances}
\label{sample_coco_longtail}
\end{wrapfigure}

\subsubsection{Datasets} 

1) LVIS~\cite{gupta2019}. It is a recent benchmark for large vocabulary long-tail instance segmentation~\cite{gupta2019}. The source images are from COCO dataset, while the annotation follows an iterative object spotting process that captures the long-tail category statistic naturally appearing in images. Current released version v0.5 contains 1,230 and 830 object classes respectively in its train and validation set, with test set unknown. 
Refer to Fig.~\ref{fig_intro} (a) for train set category distribution. 
The three sets contain about 50k, 5k and 20k images correspondingly. 
2) COCO-LT. We sample it from COCO~\cite{lin2014coco} by following an exponential distribution on training instance statistics to create a long-tail version. COCO-LT contains 80 classes and about 100k images. 
Fig.~\ref{sample_coco_longtail} shows the category distribution. Due to space limitations, we defer details of sampling process to supplement. 

\begin{table}
\centering
\caption{ Diffenrent category division scheme, with LVIS v0.5 dataset~\cite{gupta2019}. The left part is division based on training image number as in ~\cite{gupta2019}, the right part is proposed scheme based on training intance number. Train-on-val means categories that appear in the validation set}
\begin{tabular}{c|c|ccc|cccc}
\toprule
\multirow{2}{*}{Set} & \multirow{2}{*}{Total} & \multicolumn{3}{c|}{Divided by \#image} & \multicolumn{4}{c}{Divided by \#instance}            \\ \cline{3-9} 
                     &                        & \emph{rare}    & \emph{common}    & \emph{frequent}    & $(0, 10)$ & $[10, 100)$ & $[100, 1000)$ & $[1000,-]$ \\
\midrule
Train                & 1230                   & 454       & 461         & 315           & 294       & 453         & 302           & 181        \\
Train-on-val         & 830                    & 125       & 392         & 313           & 67        & 298         & 284           & 181       \\
\bottomrule
\end{tabular}
\label{data_cat_distribution}
\end{table}

\subsubsection{Metrics}
\label{eval_metric}
We adopt AP as overall evaluation metric.
Object categories in LVIS are divided into \emph{rare}, \emph{common}, \emph{frequent} sets~\cite{gupta2019}, respectively containing $<$10, 10-100, and $>=$100 training images. 
We show in Table~\ref{data_cat_distribution} the category distribution of training and validation sets. 
Besides data splitting based on image number, we devise a complimentary instance-centric category division scheme, considering number of instances is a widely adopted measurement for detection in terms of benchmark creation, model evaluation~\cite{openimage}, \cite{everingham2010pascal}, \cite{lin2014coco}. 
In particular, we divide all the  categories into four bins\footnote{Note we use ``bin'' and ``set'' interchangeably.} based on the number of training instances, with \#instances $<$10, 10-100, 100-1000, and $>=$1000, as shown in Table~\ref{data_cat_distribution}. Accordingly, we calculate AP on each bin as complementary metrics, denoted as AP$_1$, AP$_2$, AP$_3$, and AP$_4$.
Such a division scheme offers a finer dissection of model performance. For example, 
AP$_1$ corresponds to the commonly referred few-shot object detection regime~\cite{chen2018lstd}, \cite{kang2019few}, \cite{karlinsky2019repmet}. \emph{rare} set ($\leq$10 training images) contains categories that have up to 219 training instances (`chickpea'), so AP$_r$ cannot well reflect model's few-shot learning capability.
AP$_4$ reflects performance on classes with COCO level training data, while most classes in \emph{frequent} set ($>$100 images) have much less than 1,000 training instances (\textit{e.g.}, `fire-alarm': 117). 
With the two division schemes, we can report AP on both image-centric (AP$_r$, AP$_c$, AP$_f$) and instance-centric (AP$_1$, AP$_2$, AP$_3$, AP$_4$) bins for LVIS. For COCO-LT, since the per-category training instance number varies in a much larger range, we divide the categories into four bins with $<$20, 20-400, 400-8000, and $>=$8000 training instances and report performance as AP$_1$, AP$_2$, AP$_3$, AP$_4$ on these bins. 
Unless specified, AP is evaluated with COCO style by mask AP.

\subsection{Evaluating Adapted Existing Classification Methods}
\label{comp_other_methods}
We apply adapted discussed methods in Sec. \ref{other_approaches} to classification head of Mask R-CNN for long-tail instance segmentation, including ~\cite{cao2019learning}, \cite{cui2019class}, \cite{gupta2019}, \cite{lin2018focal}.
Results are summarized in Table~\ref{baseline_methods_on_cls_head}. We can see some improvements have been achieved on tail classes. 
For example, 6.0, 6.2, 7.7 absolute margins on AP$_1$ and 10.1, 8.7, 11.6  on AP$_r$ for loss re-weighting (LR), focal loss (FL) and image level repeat sampling (IS) are observed, respectively.
However, on the other hand, they inevitably lead to drop of performance on head classes, \textit{e.g.}, more than 2.0 drop for all methods on AP$_4$ and AP$_f$. Performance drop on head classes is also observed in imbalanced classification~\cite{he2009learning}, \cite{tang2008svms}. Overall AP is improved by at most 2.5 in absolute value (\textit{i.e.}, IS). Similar observation holds for box AP.

\begin{table*}
	
	\centering
	\caption{ Results on LVIS by adding common strategies in long-tail classification to Mask R-CNN in training. r50 means Mask R-CNN on ResNet50-FPN backbone with class-wise box and mask heads (standard version). CM, LR, FL and IS denote discussed class aware margin loss, loss re-weighting, Focal loss and image level repeat sampling respectively. AP$^{b}$ denotes box AP. We report result with median overall AP over 3 runs}
	\resizebox{\textwidth}{!}{
		\begin{tabular}{l|cccc|ccc|c|cccc|ccc|c}
		\toprule
		Model  & AP$_1$  & AP$_2$  & AP$_3$  & AP$_4$ & AP$_r$ &  AP$_c$ & AP$_f$  & AP & AP$_{1}^{b}$  & AP$_{2}^{b}$  & AP$_{3}^{b}$  & AP$_{4}^{b}$ & AP$_r^{b}$ & AP$_c^{b}$ & AP$_f^{b}$   & AP$^{b}$\\ 
		\midrule
		r50  & 0.0   & 17.1 & 23.7 & 29.6 &3.7& 20.0&28.4& 20.7
		& 0.0   & 15.9 & 24.6 & 30.5 & 3.3&19.0&30.0&20.8\\ 	
		CM      & 2.6   & 21.0 & 21.8 & 26.6 & 8.4 & 21.2 & 25.5 & 21.0 & 2.8 & 20.0 & 22.0 & 26.6 & 6.8 & 20.5 & 26.4& 20.7\\  
		LR      & 6.0   & 23.3 & 22.0 & 25.1 & 13.8 & 22.4 & 24.5 & 21.9 &6.0 &21.2 &22.3 &25.5 & 11.3 & 21.5 & 24.9& 21.4\\
		FL      & 6.2   & 21.0 & 22.0 & 27.0 & 12.4 &20.9 &25.9 & 21.5 & 5.8 & 20.5 & 22.7 & 28.0 & 10.5 & 21.0 & 27.0 & 21.7\\
		IS      & 7.7   & 25.6 & 21.8 & 27.4 & 15.3 & 23.7 & 25.6 & 23.2 & 6.7 & 22.8 & 22.1 & 27.4 &11.6 & 22.2 & 26.7 & 22.0\\ 
		
		%
		\bottomrule
	\end{tabular}
	}

	\label{baseline_methods_on_cls_head}
\end{table*}

\subsection{Evaluating Proposed \emph{SimCal}}
In this subsection, we report the results of our proposed method applied on mask R-CNN. We evaluate both class-wise and class-agnostic versions of the model. Here  $T$ for dual head inference is set to 300. 

\begin{table}[]
\centering
\caption{ Results on LVIS by applying \emph{SimCal} to Mask R-CNN with ResNet50-FPN. r50-ag and r50 denote models with class-agnostic and class-wise heads (box/mask) respectively. cal and dual means calibration and dual head inference. Refer to supplementary file for an anlaysis on LVIS result mean and std}
\begin{tabular}{l|c|cc|cccc|ccc|c}
\toprule
                      & Model  & cal                       & dual                      & AP$_1$                         & AP$_2$                         & AP$_3$                         & AP$_4$                         & AP$_r$ & AP$_c$ & AP$_f$ & AP       \\
\midrule
\multirow{6}{*}{bbox} & \multirow{3}{*}{r50-ag} &                           &                           & 0.0                              & 12.8                           & 22.8                           & 28.3                           & 2.8    & 16.4   & 27.5   & 18.6                                        \\
                      &  & \checkmark &                           & 12.4                           & 23.2                           & 20.6                           & 23.5                           & 17.7   & 21.2   & 23.4   & 21.5                   \\
                      &  & \checkmark & \checkmark & \textbf{12.4} & \textbf{23.4} & 21.4                           & 27.3                           & \textbf{17.7}   & \textbf{21.3}   & 26.4   & \textbf{22.7} \\
                      \cline{2-12} 
                      & \multirow{3}{*}{r50}    &                           &                           & 0.0                              & 15.9                           & 24.6 & 30.5 & 3.3    & 19.0     & 30.0     & 20.8                                           \\
                      &      & \checkmark &                           & 8.1                            & 21.0                             & 22.4                           & 25.5                           & 13.4   & 20.6   & 25.7   & 21.4                                                 \\
                      &     & \checkmark & \checkmark & \textbf{8.2}  & \textbf{21.3} & 23.0                             & 29.5                           & \textbf{13.7}   & \textbf{20.6}   & 28.7   & \textbf{22.6} \\
\midrule
\multirow{6}{*}{mask} & \multirow{3}{*}{r50-ag} &                           &                           & 0.0                              & 13.3                           & 21.4                           & 27.0                             & 2.7    & 16.8   & 25.6   & 18.0                                               \\
                      &  & \checkmark &                           & 13.2                           & 23.1                           & 20.0                             & 23.0                             & 18.2   & 21.4   & 22.2   & 21.2                                                   \\
                      &  & \checkmark & \checkmark & \textbf{13.3} & \textbf{23.2} & 20.7                           & 26.2                           & \textbf{18.2}   & \textbf{21.5}   & 24.7   & \textbf{22.2}  \\ \cline{2-12} 
                      & \multirow{3}{*}{r50}    &                           &                           & 0.0                              & 17.1                           & 23.7 & 29.6 & 3.7    & 20.0     & 28.4   & 20.7                                                  \\
                      &      & \checkmark &                           & 10.2                           & 23.5                           & 21.9                           & 25.3                           & 15.8   & 22.4   & 24.6   & 22.3                                           \\
                      &     & \checkmark & \checkmark & \textbf{10.2} & \textbf{23.9} & 22.5                           & 28.7                           & \textbf{16.4}   & \textbf{22.5}   & 27.2   & \textbf{23.4} \\
\bottomrule
\end{tabular}
\label{main_results_lvis_table}
\end{table}

\begin{table*}
\renewcommand{\tabcolsep}{0.8pt}
\begin{floatrow}
\capbtabbox{
	\begin{tabular}{l|cccc|ccc|c}
		\toprule
		Model  & AP$_1$  & AP$_2$  & AP$_3$  & AP$_4$ & AP$_r$ & AP$_c$  &  AP$_f$  & AP\\ 
		\midrule
		r50  & 0.0   & 17.1 & 23.7 & 29.6 &3.7& 20.0&28.4& 20.7\\ 
		CM      & 4.6 & 21.0 & 22.3 & 28.4 & 10.0 & 21.1 & 27.0 & 21.6\\ 
		LR      & 6.9 & 23.0 & 22.1 & 28.8 & 13.4 & 21.7 & 26.9 & 22.5 \\
		FL      & 7.1 & 21.0 & 22.1 & 28.4 & 13.1 & 21.5 & 26.5 & 22.2 \\
		
		IS      & 6.8 & 23.2 & 22.5 & 28.0 & 14.0 & 22.0 & 27.0 & 22.7\\ 
		
		ours  & \textbf{10.2}  &	\textbf{23.9} &	22.5& 28.7 & \textbf{16.4} & \textbf{22.5} & 27.2& \textbf{23.4} \\ 
		\bottomrule
	\end{tabular}
}{
	\caption{ Results for augmenting discussed long-tail classification methods with proposed decoupled learning and dual head inference.}
	\label{decoupled_long_tail_classification}
}

\capbtabbox{
	\begin{tabular}{l|cccc|c}
		\toprule
		& AP$_1$  & AP$_2$  & AP$_3$  & AP$_4$   & AP\\ 
		\midrule
		orig      & 0.0   & 13.3 & 21.4 & 27.0 & 18.0 \\ 
		cal     & 8.5   & 20.8 & 17.6 & 19.3 & 18.4 \\ 
		avg      & 8.5   & 20.9 & 19.6 & 24.6 & 20.3\\ 
		sel     & \textbf{8.6}   & \textbf{22.0} & 19.6 & 26.6 & \textbf{21.1}\\ 
		\bottomrule
	\end{tabular}
}
{
	\caption{ Comparison between proposed combining scheme (sel) and averaging (avg).}
	\label{combination_method_ablation}
}

\end{floatrow}
\end{table*}

\subsubsection{Calibration Improves Tail Performance} From results in Table \ref{main_results_lvis_table}, we observe consistent improvements on tail classes for both class-agnostic and class-wise version of Mask R-CNN (more than 10 absolute mask and box AP improvement on tail bins). Overall mask and box AP are boosted by a large margin. But we also observe a significant drop of performance on head class bins, \textit{e.g.}, 23.7 to 21.9 on AP$_3$ and 29.6 to 25.3 on AP$_4$ for the class-wise version of Mask R-CNN. With calibration, the classification head is effectively balanced.

\subsubsection{Dual Head Inference Mitigates Performance Drop on Head Classes}  The model has a minor performance drop on the head class bins but an enormous boost on the tail class bins. For instance, we observe 0.8 drop of AP$_4$ but 13.3 increase on AP$_1$ for r50-ag model. It can be seen that with the proposed combination method, the detection model can effectively gain the advantage of both calibrated and original classification heads.

\subsubsection{Class-wise Prediction is Better for Head Classes While Class-agnostic One is Better for Tail Classes} 
We observe AP$_1$ of r50-ag with cal and dual is 3.1 higher (13.3 vs 10.2) than that of r50 while AP$_4$ is 2.5 lower (26.2 vs 28.7), which means class-agnostic heads (box/mask) have an advantage on tail classes, while class-wise heads perform better for many-shot classes.
This phenomenon suggests that a further improvement can be achieved by using class-agnostic head for tail classes so they can benefit from other categories for box and mask prediction, and class-wise head for many-shot classes as they have abundant training data to learn class-wise prediction, which is left for future work.

\subsubsection{Comparing with Adapted Existing Methods}
For fair comparison, we also consider augmenting the discussed imbalance classification approaches with proposed decoupled learning framework.
With the same baseline Mask R-CNN trained in the normal setting, we freeze other parts except for classification head, and use these methods to calibrate the head. 
After that, we apply the dual head inference for evaluation. 
As shown in Table~\ref{decoupled_long_tail_classification}, they have similar performance on head classes as dual head inference is used. They nearly all get improved on tail classes than the results in Table~\ref{baseline_methods_on_cls_head} (\textit{e.g.}, 4.6 vs 2.6, 6.9 vs 6.0, and 7.1 vs 6.2 on AP$_1$ for CM, LR, and FL methods respectively), indicating the effectiveness of the decoupled learning scheme for recognition of tail classes. The image level repeat sampling (IS) gets worse performance than that in Table~\ref{baseline_methods_on_cls_head}, suggesting box and mask learning also benefits a lot from the sampling. Our method achieves higher performance, \textit{i.e.}, 10.2 and 23.9 for AP$_1$ and AP$_2$, which validates effectiveness of the proposed bi-level sampling scheme.

\begin{figure}
	\centering
	\includegraphics[width=\linewidth]{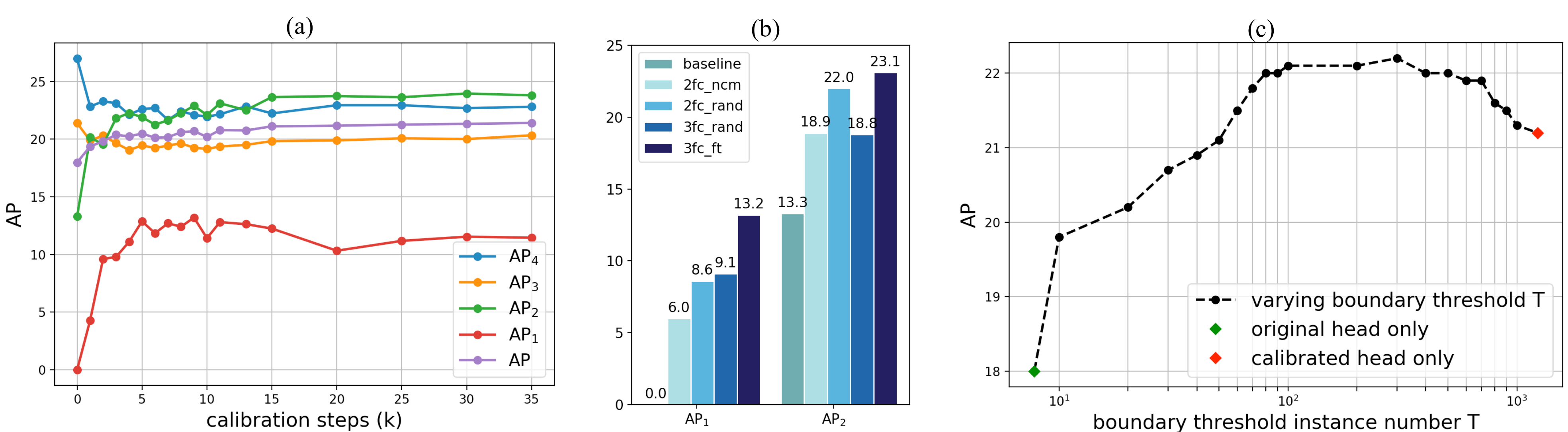}
	\caption{ (a) Model performance as a function of calibration steps. 
	The result is obtained with r50-ag model (Table~\ref{main_results_lvis_table}). 
	(b) Effect of design choice of calibration head. Baseline: original model result; 2fc\_ncm~\cite{guerriero2018deepncm}: we have tried to adopt the deep nearest class mean classifier learned with 2fc representation. 2fc\_rand: 2-layer fully connected head with 1024 hidden units, random initialized; 3fc-rand: 3-layer fully connected head with 1024 hidden units, random initialized. 3fc-ft: 3fc initialized from original head.
	(c) Effect of boundary number $T$ (with r50-ag)}
	\label{ablation_cal_steps_and_head_choices_and_threshold}
\end{figure}

\subsection{Model Design Analysis of \emph{SimCal}}
\label{ablation_experiments}
\subsubsection{Calibration Dynamics}
As shown in Fig.~\ref{ablation_cal_steps_and_head_choices_and_threshold} (a), with the progress of calibration, model performance is progressively balanced over all the class bins. Increase of AP on tail bins (\textit{i.e.}, AP$_1$, AP$_2$) and decrease of AP on the head (\textit{i.e.}, AP$_3$, AP$_4$) are observed. With about 10-20k steps, AP on all the bins and overall AP converge to a steady value.

\subsubsection{Design Choice of Calibration Head}
\label{design_choice_calibration_head}
While the proposed calibration method tries to calibrate the original head, we can perform the calibration training on other head choices. 
As shown in Fig.~\ref{ablation_cal_steps_and_head_choices_and_threshold} (b), we have tried different instantiations instead of the original head.
It is interesting that with random initialization, 3-layer fully connected head performs worse than 2-layer head on AP$_1$ (\textit{i.e.}, 2fc\_rand vs 3fc-rand). But when it is initialized from the original 3-layer head, the performance is significantly boosted by 4.1 and 4.3 AP respectively on AP$_1$ and AP$_2$ (\textit{i.e.}, 3fc\_ft). This phenomenon indicates that training under random sampling can help the classification head learn general features and perform well when calibrating with balanced sampling. We only compare them on the tail class bins since they perform on par on head class bins with dual head inference.

\subsubsection{Combining Scheme and Head/Tail Boundary for Dual Heads}
\label{heads_combination_ablation}
As shown in Table~\ref{combination_method_ablation}. Our combination approach achieves much higher performance than simple averaging. Refer to supplementary material for more alternative combining choices. We also examine the effect of head/tail boundary as in Fig.~\ref{ablation_cal_steps_and_head_choices_and_threshold} (c). For the same model, we vary the boundary threshold instance number $T$ from 10 to 1000. The AP is very close to optimal ($T=300$) in $T\in [90, 500]$. Thus dual head is insensitive to the exact value of hyperparameter T in a wide range.

\begin{table}[]
\centering

\caption{ Results with Hybrid Task Cascade (HTC) on LVIS. With backbone of ResNeXt101-64x4d-FPN. best denotes best single model performance reported in ~\cite{gupta2019}. The remaining rows are our experiment results with HTC. 2fc\_rand and 3fc\_ft are different design choices of classification head (Sec.~\ref{design_choice_calibration_head}). Only 2fc\_rand is available on test set as the evaluation server is closed}
\begin{tabular}{c|cccc|ccc|c|ccc|c}
\toprule
\multirow{2}{*}{Model} & \multicolumn{8}{c|}{Val}                                                                          & \multicolumn{4}{c}{Test}       \\ \cline{2-13} 
                       & AP$_1$        & AP$_2$        & AP$_3$ & AP$_4$ & AP$_r$        & AP$_c$ & AP$_f$ & AP            & AP$_r$ & AP$_c$ & AP$_f$ & AP   \\ \midrule
best~\cite{gupta2019}  & --            & --            & --     & --     & 15.6          & 27.5   & 31.4   & 27.1          & 9.8    & 21.1   & 30.0   & 20.5 \\ \hline
htc-x101               & 5.6           & 33.0          & 33.7   & 37.0   & 13.7          & 34.0   & 36.6   & 31.9          & 5.9    & 25.7   & 35.3   & 22.9 \\
IS    & 10.2          & 32.3          & 33.2   & 36.6   & 17.6          & 33.0   & 36.1   & 31.9          & --     & --     & --     & --   \\
2fc\_rand               & 12.9          & 32.2          & 33.5   & 37.1     & 18.5          & 33.3   & 36.1     & 32.1          & 10.3   & 25.3   & 35.1   & 23.9 \\
3fc\_ft                 & \textbf{18.8} & \textbf{34.9} & 33.0   & 36.7   & \textbf{24.7} & 33.7   & 36.4   & \textbf{33.4} & --     & --     & --     & --   \\ \bottomrule
\end{tabular}
\label{htc_results}
\end{table}

\begin{table*}[]
	
	\centering
	\caption{ Results on COCO-LT, evaluated on minival set.  AP$_1$, AP$_2$, AP$_3$, AP$_4$ correspond to bins of [1, 20), [20, 400), [400, 8000), [8000, -) training instances}
	\begin{tabular}{lcc|cccc|c|cccc|c}
		\toprule
		Model   & cal & dual & AP$_{1}$  & AP$_{2}$  & AP$_{3}$  & AP$_{4}$   & AP & AP$_{1}^{b}$  & AP$_{2}^{b}$  & AP$_{3}^{b}$  & AP$_{4}^{b}$   & AP$^{b}$\\ 
		\midrule
		r50-ag   & & & 0.0   & 8.2 & 24.4 & 26.0 & 18.7
		& 0.0   & 9.5 & 27.5 & 30.3 & 21.4\\ 
		r50-ag   & \checkmark &  & 15.0	& 16.2 & 22.4 &	24.1 & 20.6
		& 14.5   & 17.9 & 24.8 & 27.6 & 22.9\\ 
		r50-ag  & \checkmark &  \checkmark  & \textbf{15.0}	& \textbf{16.2} & 24.3 &	26.0 & \textbf{21.8}
		& \textbf{14.5}   & \textbf{18.0} & 27.3 & 30.3 & \textbf{24.6}\\ 
		\midrule
	\end{tabular}
	\label{result_coco_lt_kd}
\end{table*}



\subsection{Generalizability Test of \emph{SimCal}}
\subsubsection{Performance on SOTA Models}
We further apply the proposed method to state-of-the-art multi-stage cascaded instance segmentation model, Hybrid Task Cascade~\cite{chen2019hybrid} (HTC), by calibrating classification heads at all the stages. As shown in Table~\ref{htc_results}, our method brings significant gain on tail classes and minor drop on many-shot classes. Notably, the proposed approach leads to much higher gain than the image level repeat sampling method (IS), (\textit{i.e.}, 8.5 and 2.5 higher on AP$_1$ and AP$_2$ respectively). We achieve state-of-the-art single model performance on LVIS, which is 6.3 higher in absolute value than the best single model reported in~\cite{gupta2019} (33.4 vs 27.1). And with test set, a consistent gain is observed.

\subsubsection{Performance on COCO-LT}
\label{coco_lt_experiments}

As shown in Table~\ref{result_coco_lt_kd}, 
similar trend of performance boost as LVIS dataset is observed.
On COCO-LT dual head inference can enjoy nearly full advantages of both the calibrated classifier on tail classes and the original one on many shot classes. But larger drop of performance on many-shot classes with LVIS is observed. It may be caused by the much stronger inter-class competition as LVIS has much larger vocabulary.

\section{Conclusions}
In this work, we carefully investigate two-stage instance segmentation model's performance drop with long-tail distribution data and reveal that the devil is in proposal classification. 
Based on this finding, we first try to adopt several common strategies in long-tail classification to improve the baseline model. We also propose a simple calibration approach, \emph{SimCal}, for improving the second-stage classifier on tail classes.
It is demonstrated that \emph{SimCal} significantly enhances Mask R-CNN and SOTA multi-stage model HTC. 
A large room of improvement still exists along this direction.  
We hope our pilot experiments and in-depth analysis together with the simple method would benefit future research.

\subsubsection{Acknowledgement} Jiashi Feng was partially supported by MOE Tier 2 MOE 2017-T2-2-151, NUS\_ECRA\_FY17\_P08, AISG-100E-2019-035.

\renewcommand{\thesection}{\Alph{section}}

\section{Implementation Details}

We use PyTorch to implement the proposed method. Our implementation is based
on the mmdetection~\cite{mmdetection}. All the models are trained with SGD and momentum of 0.9, on 8 NVIDIA Tesla V100 GPUs.

\subsection{Standard Model Training}
For standard model training, i.e., normal training of whole model with random sampling as shown in Fig.2 (a), 8 images per mini-batch are sampled. 

For the Mask-RCNN standard model training, the learning rate starts with 0.01 and decays at the 8th and 11th epochs by a factor of $0.1$. The training ends at the 12th epoch. The short edge of images is fixed at 800 pixels and the long edge is capped at 1,333 pixels, without changing the aspect ratio. 

For the HTC standard model training, the learning rate starts with 0.01 and decays at the 16th and 19th epochs by a factor of $0.1$. The training ends at the 20th epoch. We also add multi-scale augmentation for HTC model training. More specifically, the size of image short edge is randomly sampled from [400, 1,400], and the long edge is capped at  1,600 pixels, without changing the aspect ratio. 

\subsection{Calibration Training}
For calibration training, we sample 16 classes and 1 image per class in one mini-batch.  proposals are matched to ground truth bounding boxes with threshold of 0.5, as in~\cite{ren2015faster}. We  sample the same number of background ROIs as the  the foreground ROIs (i.e., \# background : \# foreground = 1:1). For both Mask-RCNN and HTC calibration, the learning rate starts with 0.01 and decays at the 8000th (i.e., 0.001) and 11000th  (i.e., 0.0001) steps by a factor of 0.1. The calibration training ends at 12000 steps.

\subsection{Other Hyperparameters}
All the hyperparameters for the adopted long-tail classification methods are tuned  by grid search. (i) For re-weighting method, $N$ (i.e., the numerator of class dependent loss weight, as in Sec. 4.1) and background loss weight   are set as 100  and 1 respectively. (ii) For Focal loss, $\gamma$ and $\alpha$  are set as 3 and 0.5 respectively. Different from the observation made in~\cite{lin2018focal} that one-stage detector's performance is relatively robust to  the  value of $\gamma$ in a wide range, we find the performance of Mask R-CNN model on LVIS is sensitive to the value of $\gamma$ and $\gamma=3$  gives the best performance. The hyperparameter $C$ for class-aware margin loss is set as 6.0.

\section{Qualitative Results}

\begin{figure*}
	\centering
	\includegraphics[width=1.0\textwidth]{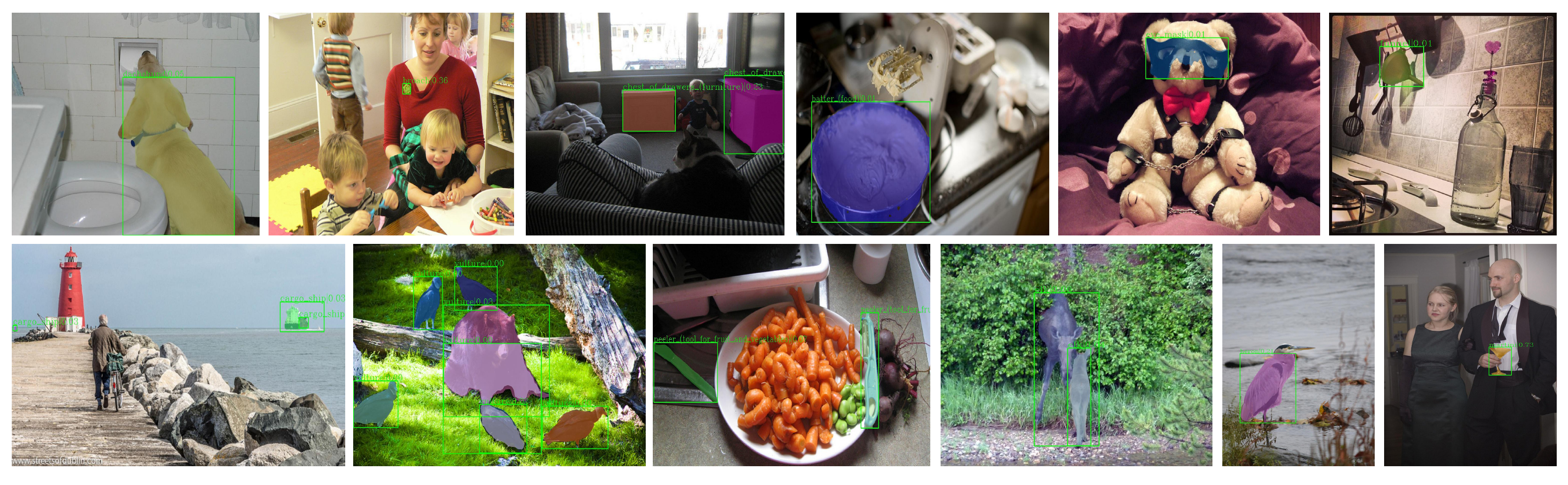}
	\caption{ Qualitative results for low-shot categories (in [1, 10) bin) on LVIS with r50-ag model. Only relevant detections of low-shot classes are visualized for a better view. Although there are some false positives, the model after calibration can detect and segment those object instances. For the original model without calibration, all those objects are missed. Note some detections have 0.00 score as we only round the score to 2 decimal places}
	\label{qualitative_result}
\end{figure*}

Some qualitative results can be found in Fig.~\ref{qualitative_result}.

\section{COCO-LT Sampling}
{Here we explain how we created the COCO-LT dataset. }
We evenly divide the 80 categories into 4 subsets according to their   category index, i.e.  (1-20, 21-40, 41-60, 61-80) respectively, each with 20 classes. For the $i$th subset if $i \neq 1$, we  randomly sample $n_i$ instances with $n_i \in   (8*10^{4-i}, 8*10^{5-i})$. For the first subset (with category indices of 1-20, i.e., i=1) we do not perform sampling. If an instance is not sampled, we remove it from the annotation of its image. For training images without sampled instances, we remove these images. The category distribution after sampling (COCO-LT as shown as in Fig. 3) follows a long-tail distribution. The validation set is kept as the original and   used for evaluation.

%

\section{How to Combine the Dual Heads}
In addition to the proposed simple threshold selection scheme for combining the calibrated and original heads' prediction, we also explored some other possible schemes.
As shown in Table~\ref{combination_method_ablation}, we compare the proposed combination scheme with  other  alternatives,  including  (i) \emph{cal-only}: using only the prediction from the calibrated head; (ii) \emph{avg}: averaging predictions of the original head and calibrated head, this is widely adopted way of ensembling two classification models; 
(iii) \emph{det}: using the two heads separately for detection outputs and combining them afterward (i.e., with NMS), this is most simple but effective way of ensembling detection models; 
(iv) \emph{sel}: the proposed output combining scheme; 
(v) \emph{sel-thr}: filtering the calibrated head predictions with 0.05 threshold before \emph{sel}, aiming to reduce low quality detections from calibrated head with low confidence score; (vi) \emph{sel-scale}: scaling calibrated head's predictions by ratio of average background score between calibrated and original head's predictions before \emph{sel}, since the calibrated head is trained with different background and foreground sample ratio, it has different average foreground score compared to original head, \emph{sel-scale} aims to scale alleviate the difference; 
(vii) \emph{sel-norm}: normalizing the prediction by the summed score over classes after \emph{sel}, aiming to convert the prediction to a normalized classification score.

The proposed combining scheme achieves the best overall result (i.e., 21.1 AP) compared with the other alternatives. Those prior strategies of \emph{sel-thr}, \emph{sel-scale} and \emph{sel-norm} all have similar but slightly lower performance. The result verifies the simplicity and effectiveness of proposed combining method.

\begin{table}[h!]
	\small
	\centering
	\renewcommand{\tabcolsep}{2.5pt}
	\renewcommand{\arraystretch}{1.1}
	\caption{Ablation result for different ways of combining calibrated and original heads' predictions. The model is Mask R-CNN with ResNet50-FPN backbone and class agnostic box and mask heads. The experiment is with 2 layer fully connected head with random initialization (i.e., 2fc)}
	\begin{tabular}{l|cccc|c}
		\toprule
		Model   & AP$_1$  & AP$_2$  & AP$_3$  & AP$_4$   & AP \\ 
		\midrule
		\emph{orig}      & 0.0   & 13.3 & \textbf{21.4} & \textbf{27.0} & 18.0\\ 
		\emph{cal-only}     & 8.5   & 20.8 & 17.6 & 19.3 & 18.4\\ 
		\emph{avg}      & 8.5   & 20.9 & 19.6 & 24.6 & 20.3 \\ 
		\emph{det}      & 8.6   & 22.0 & 16.7 & 25.2 & 19.8 \\ 
		\emph{sel}      & \textbf{8.6}   & \textbf{22.0} & 19.6 & 26.6 & \textbf{21.1}\\ 
		\emph{sel-thr}      & 8.5   & 20.8 & 20.1 & 26.7 & 20.9\\ 
		\emph{sel-scale}      & 8.5   & 21.3 & 19.9 & 26.7 & 21.0\\  
		\emph{sel-norm}      & 8.5   & 21.9 & 19.5 & 26.2 & 20.9\\
		\bottomrule
	\end{tabular}
	\label{combination_method_ablation}
	
\end{table}

\section{How Calibration Learning Rate Affects Performance}
While we use the  same initial learning rate for calibration as standard model training (i.e., starting from 0.01), we examine the results when varying the initial learning rate for calibration, to see if optimal learning rate for calibration is different from standard model training. We measure model performance with overall AP. As shown in Table~\ref{lr_ablation}, we tested the following learning rates of 0.001, 0.002, 0.004, 0.008, 0.01, 0.02, 0.04 and 0.08. The decaying step and factor remains the same. The best performance is achieved at learning rate of 0.01, same as standard model training. The phenomenon also stands in contrast to the observation in conventional fine-tuning that a much lower learning rate compared to pre-training is required for optimal performance.


\begin{table}[!h]
	\small
	\centering
	\renewcommand{\tabcolsep}{3.5pt}
	\renewcommand{\arraystretch}{1.1}
	\caption{Ablation study for calibration learning rate. The model is Mask R-CNN with ResNet50-FPN backbone and class agnostic box and mask heads}
	\begin{tabular}{lcccccccc|c}
		\toprule
		lr   & $0.001$  & $0.002$  & $0.004$  & $0.008$ & $0.01$  & $0.02$ & $0.04$ & $0.08$ & baseline\\
		\midrule
		AP   &19.5   	 & 21.8& 21.9& 22.0& \textbf{22.2}& 21.5 & 21.1 & 21.0 &18.0 \\ 
		\bottomrule
	\end{tabular}
	\label{lr_ablation}
\end{table}

\section{Layers to Perform Calibration}
While we calibrate the whole 3-layer fully connected classification head, we tried to only perform the calibration on last layer, and last 2 layers, to see if we can achieve better performance. We measure model performance with overall AP. Table~\ref{cal_layer_ablation} shows the result comparison of those settings. Best result is obtained when we calibrate the whole classification head.

\begin{table}[h!]
	\small
	\centering
	\renewcommand{\tabcolsep}{3.5pt}
	\renewcommand{\arraystretch}{1.1}
	\caption{Ablation study for calibration layers. lastfc means only calibrate the last fc layer of classification layerl; last2fc indicates calibrating the last 2fc layers; all 3fc means normal setting of calibrating the full 3 layer classification head. The model is Mask R-CNN with ResNet50-FPN backbone and class agnostic box and mask heads}
	\begin{tabular}{lccc|c}
		\toprule
		layer   & $lastfc$  & $last2fc$  & $all 3fc$ &baseline\\
		\midrule
		AP   &21.9 & 21.8& \textbf{22.2} & 18.0\\
		\bottomrule
	\end{tabular}
	\label{cal_layer_ablation}
\end{table}

\section{LVIS Mean and Std Analysis}
As shown in Tab~\ref{mean_std_results}, we report mean and std analysis for the major two \emph{SimCal} models on LVIS dataset.  As shown from the results, the variance of AP result is much larger for the tail classes while smaller for many-shot classes as they have ample training instances.

\begin{table}[h!]
	\small
	\centering
	\renewcommand{\tabcolsep}{2pt}
	\renewcommand{\arraystretch}{1.1}
	\caption{Mean and standard deviation analysis for \emph{SimCal} models on LVIS dataset. The result is obtained by repeat each experiments 5 times and calculating the mean and standard deviation of each metric}
	\begin{tabular}{l|cccc|ccc|c}
		\toprule
		Model  & AP$_1$  & AP$_2$  & AP$_3$  & AP$_4$ & AP$_r$ & AP$_c$  &  AP$_f$  & AP\\ 
		\midrule
		r50-ag-lvis  & 13.0$_{\pm1.5}$ & 23.2$_{\pm0.8}$ & 20.7$_{\pm0.3}$       & 26.2$_{\pm0.2}$     & 18.0$_{\pm0.9}$   & 21.3$_{\pm0.3}$  & 24.8$_{\pm0.1}$   & 22.4$_{\pm0.3}$ \\
		r50-lvis      &  10.2$_{\pm1.3}$ & 23.9$_{\pm0.6}$ & 22.5$_{\pm0.4}$   & 28.7$_{\pm0.1}$  & 16.4$_{\pm0.7}$   & 22.5$_{\pm0.4}$  & 27.2$_{\pm0.2}$   & 23.4$_{\pm0.2}$ \\ 
		\bottomrule
	\end{tabular}
	\label{mean_std_results}
\end{table}


\clearpage
\bibliographystyle{splncs04}
\bibliography{egbib}
\end{document}